# Coupling Phonology and Phonetics in a Constraint−Based Gestural Model

Markus Walther[1] and Bernd J. Kröger[2]

[1] Seminar für Allgemeine Sprachwissenschaft, Heinrich-Heine-Universität Düsseldorf
Universitätsstr. 1, 40225 Düsseldorf
[2] Institut für Phonetik der Universität zu Köln
Greinstr.2, 50939 Köln

email: walther@sapir.ling.uni-duesseldorf.de, amp04@rs1.rrz.uni-koeln.de

**Abstract.** An implemented approach which couples a constraint-based phonology component with an articulatory speech synthesizer is proposed. Articulatory gestures ensure a tight connection between both components, as they comprise both physical-phonetic and phonological aspects. The phonological modelling of e.g. syllabification and phonological processes such as German final devoicing is expressed in the constraint logic programming language CUF. Extending CUF by arithmetic constraints allows the simultaneous description of both phonology and phonetics. Thus declarative lexicalist theories of grammar such as HPSG may be enriched up to the level of detailed phonetic realisation. Initial acoustic demonstrations show that our approach is in principle capable of synthesizing full utterances in a linguistically motivated fashion.

## 1 Introduction

Within the last few years it has been stressed in the field of machine speech recognition and synthesis that one should incorporate genuine linguistic information, especially of phonological nature, into models and applications. Researchers expect such integration of linguistic insights to be profitable for dimensions such as robustness, recognition rates, naturalness of pronunciation etc. On the other hand the so-called 'laboratory' branch of phonology stresses the necessity of obtaining detailed phonetic foundation to support phonological theories (Kingston 1990). Although the question of how to realize abstract phonological information phonetically thus has gained momentum, there are nevertheless few models coupling both domains, which are at the same time linguistically satisfying. This paper describes such an approach to realize a speech production model.

A crucial feature of this model is that it employs a computer–implemented constraint–based conception of phonology, thus being amenable to embedding into declarative formal theories of grammar such as HPSG (Pollard & Sag 1987). It follows that it is possible in principle to give a parallel description of the contribution of different linguistically relevant dimensions (e.g. morphological word accent, semantic focus, syntactic sentence modus) to detailed phonetic realisation.

A characteristic of our approach is the absence of any explicit interface between phonological and phonetic *description*. This follows from the bivalent character of *articulatory gestures* as our fundamental concept: on the one hand they are abstract phonological entities[3], on the other hand every gesture is realized as a target-directed movement of a specific articulator (e.g. to effect lip closure), having both a concretely defined temporal extension and relationships to other gestures. In our approach we specify e.g. the temporal relations between gestures[4] in the same formal description as, for example, phonological alternations.

Gestures as concretized in the above fashion immediately form the input to an articulatory speech synthesis component. Its goal is to model the human speech production mechanisms as naturally as possible. The phonetic production model employed comprises both a dynamic model whose task is to generate articulator movements on the basis of quantitatively and temporally fully specified gestures and an articulatory–acoustic model, i.e. an artificial vocal tract. The latter is responsible for the generation of acoustic speech output, which finally serves to make the abstract linguistic input audible.

## 2   The dynamic model and articulatory gestures

The dynamic model is responsible for computing the cinematics of articulatory movements from gestures.

Phonetically, gestures correspond to spatially goal–directed movements which can be temporally localised. They realize an intentional and linguistically relevant constriction (Saltzman und Munhall 1989; Browman & Goldstein 1992). The goal can be quantified as a concrete target coordinate in the control parameter space of the corresponding articulator. Each gesture influences the articulator realizing it only within a defined time interval, the gestural activation interval. If no gesture is active for a given articulator, it approaches its inherent neutral position. The neutral position of all articulators in our model corresponds to the production of a non–nasal schwa. In physical terms we describe the motion of a gesturally activated articulator by a critically damped harmonic oscillator.[5] For every gesture one can define an inherent time scale, the *phase scale*. Phase values parametrize the relative articulator–target distance, i.e. to what degree a gesture has been executed: for a given phase value we always find the same relative articulator–target distance Kröger 1993a, 1993b). The inherent gestural parameter *eigenperiod* denotes the absolute length of the temporal interval given by 0 to 360 degree, thereby defining the temporal stretchedness of a gestural phase scale. Eigenperiods differ between vocalic, consonantal and opening, i.e. glottal and velic gestures. Decreasing eigenperiod corresponds to a faster execution of the respective gesture: a constant relative articulator–target distance is reached more quickly.

---

[3] Gestures do *not* correspond to the traditional category 'phoneme'. Determining their mutual relationship is a problematic issue – in a rough approximation we may map one or more gestures onto a phoneme.

[4] As we shall see, these relations are mainly determined by which syllable functions are associated with them.

Gestures are temporally coordinated within an utterance in a relative gesture–by–gesture fashion. For this the gestural parameter of *association phase* is necessary. The abstract point in time corresponding to the association phase of one gesture is equated with a similar phase point of the other gesture.[6] A detailed description of these phasing rules is given by Kröger (1993b). Basically, a first step consists of the gap-free concatenation of all vocalic gestures. The extremal points of such vocalic activation intervals now define anchor points for the required coordination with consonantal closing gestures. Thus there is always overlap between consonantal and vocalic gestures, and the coproduction that results automatically models aspects of coarticulation. Finally, a gestural parameter *release phase* determines the abstract end of the activation interval, being context-sensitive like its association phase counterpart.

## 3   The articulatory–acoustic model

The articulation model of our articulatory–acoustic component is responsible for computing the midsagittal contour from its input articulatory movements, from which a 3-D geometry of the vocal tract is formed in turn. Additionally it computes the instantaneous glottal opening area from control parameters *vocal fold tension* and *glottal opening degree* (i.e. mean vocal fold distance, the articulatory component of vocal fold distance). From these geometrical data we are then able to compute the speech signal by means of the acoustic model. The articulatory model defines the inventory of model articulators and the range of corresponding control parameter values. As the acoustics of the speech output is determined solely from vocal tract geometry – apart from phonatory contribution –, control parameters have been defined in such a way as to directly describe constriction degree and location of their corresponding articulators.

The acoustic model comprises a vocal tract model and a self–oscillating glottis model. The vocal tract model is responsible for computing the pressure and airflow conditions within the vocal tract for each location and each time instant.

---

[5] For illustration one may imagine a mass–spring system immersed in a highly viscous medium trying to oscillate when deflected from a rest position.

[6] Witness the computation of the temporal endpoint of a gesture in (6).

Friction noise is generated automatically upon fulfillment of relevant aerodynamic conditions responsible for turbulent flow.[7] The self–oscillating glottis model is controlled by parameters *vocal fold tension* and *degree of glottal opening*, the latter denoting the actively controllable articulatory tuning of the vocal folds. Thus oscillation results only if a suitable tension of the vocal folds is met by suitable subglottal pressure.

## 4   The phonological component

The phonological component is responsible for modelling those parts of the sound-related linguistic subsystem, which are language-specific and not determined by physics and biology. The base paradigm we want to employ is that of Declarative Phonology (DP, cf. Scobbie 1991), which emphasizes that only wellformedness conditions for a solution are to be stated, but not the steps to reach that solution (declarativity). Note that this contrasts with previous generative approaches characterized by a procedural flavour. DP dispenses with extrinsic rule ordering, formally equates phonological rules and representations, makes feature-changing impossible (monotonicity) and gets rid of levels of derivation (monostratality). Phonological descriptions are distinguished from their denotation in the world, i.e. classes of sound events.

Drawing on Bird(1991:143) we require phonological descriptions to possess formal adequacy in that they have to be expressed in a description language for which a formal syntax and semantics exists. The description language we use here is CUF (Dörre & Eisele 1991). CUF is a constraint logic programming language designed with linguistic applications in mind, allowing for example HPSG grammars to be implemented. Preparatory groundwork on declarative syllabification and the formalisation of phonological processes has been documented in (Walther 1992, 1993 ; Walther & Wiese 1993).

Let us now illustrate the general approach, taking the German verbal form **ebbt** *to ebb, 3sg pres.ind.* as our example. As we can see from **ebbt** – [ʔɛptʰ] vs. **ebben** – [ʔɛbən], the labial plosive undergoes a contextually determined voicing alternation ([p] vs. [b]). This is an instance of the process of German final devoicing, which can be described by the following two surface-true generalisations, which are taken to be rather uncontroversial with respect to the literature, expressed as (1).a and (2).a with examples as b., respectively.

(1)   a.   Obstruents are voiceless in the syllable coda (`coda`).
      b.   Jagd [jaːkt.], Bund [bʊnt.], Lob [loːp.], ebbt [ɛpt.], brav [bʁaːf.]

(2)   a.   Alternating obstruents are voiced in syllable onset (`in_onset`)
      b.   jagen [jaːgən], Bundes [bʊn.dəs], Lobes [loː.bəs], Ebbe [ɛb͡ə],
           braver [bʁaː.vɐ]

---

[7] I.e. sufficiently narrow constriction in the vocal tract coupled with sufficiently high airflow.

In the following we give an exposition of the steps leading to a formal modelling of the example at hand. First of all we specify segments (`seg`) as units consisting of primary and secondary gestures. In CUF: `segment :: primary:gesture, secondary:gesture`. Phonologically we may conceive of the gesture as an underlying abstract unit (Browman & Goldstein 1989, 1992), defining discrete categories such as articulator, articulator position and constriction degree. Categories for articulators or positions would be e.g. *lips, tongue tip, vocal folds* or *alveolar, postalveolar, palatal, velar*. Categories for manner of gestural articulation would be e.g. full closure, near closure or approaching closure, corresponding to the realisation of a plosive, fricative or vocalic constriction.

Next we need to tie segmental positions to syllable structure. For this purpose a suitable type hierarchy, as familar e.g. from the HPSG literature, is made available, cf. (3).

(3)

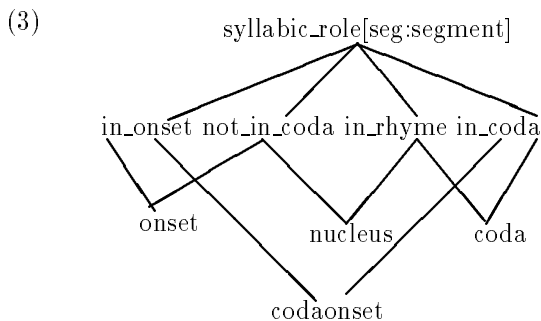

By decomposing subsyllabic roles as in (3), we are able to represent e.g. a syllable nucleus as simultaneously being in rhyme and not in the syllable coda, in agreement with phonological intuition. Remarkably, ambisyllabic positions (`codaonset`), as needed e.g. for **Ebbe**, can be characterised by static type description alone, thereby dispensing with the need to describe them by means of configurational dominance.

In a generative approach generalisations (1) and (2) have been described by means of a *feature changing rule*, which operated over underlyingly voiced representations for the alternating obstruents, changing their voice feature. Here (1) and (2) will be reformulated as *monotonic* conditional constraints (4) and (5). We display a predicate logic formalisation side by side with its CUF equivalent, the latter being derived using wellknown logical equivalences and employing the logical connectives negation ($\sim$), disjunction (;) and conjunction (&) to form composite formulae out of attribute:value descriptions.

(4)  a.  $\forall\, segment\, S : obstruent(S) \rightarrow (coda(S) \rightarrow voiceless(S))$

b.  `final_devoicing := self:(seg:~obstruent ; seg:obstruent & (~coda ; coda & seg:secondary:voiceless)).`

(5)  a.  $in\_onset(S) \rightarrow voiced(S)$

b.  `voiced_in_onset := self:(~in_onset ; in_onset & seg:secondary:inactive).`

We can conceive of (4) as a lexical constraint on the segmental inventory, whereas (5) is part of enriched segmental representations for the class of alternating obstruents (/b,d,g,v,z,ʒ/). Conforming to the lexicalist approach of HPSG these final devoicing *representations* thus are part of regular lexicon entries.

How then does phonetics enter into abstract phonological representations? In this model we assume that *phonological descriptions are underspecified phonetic descriptions*. In accordance with this view we simply add additional information, which predominantly reflects the quantitative dimensions of gestures:
`gesture :: cd:constriction_degree, cl:constriction_location,`
`clip`[8]`:clipping, time:timing. timing :: start:number, end:number,`
`eigenperiod:number, assoc:number, release:number.` Apart from constriction degree and location, which where already present in phonology, in particular information for eigenperiod, association and release phase is added.[9] These control parameters are filled with concrete gesture-specific values taken from Kröger (1993a,b). Next we need to instantiate the absolute temporal placement of gestures for a given utterance, which depends on the syllable context and is not lexically specified. For this task a suitable adaption of the association rules[10] given by Kröger (1993b) is used, specifying intergestural timing within CUF by means of simple recursive definitions.

For a general treatment of the temporal domain we found it necessary to extend CUF by *arithmetic constraints*. This extension allows us to formulate (and solve practically) equations containing the usual arithmetic operations over the domain of reals, thereby adding expressive power to the feature term constraints already offered by the language. Thus we are able to compute e.g. the end time of a gesture by $endtime_{gesture} = start_{gesture} + eigenperiod * release_{gesture}/360$, even if its start time is *unknown* at the time of computation. This is an important property which ensures that the order-free characteristics of CUF are preserved. We formulate the above equation in CUF as in (6).[11] Next to it the full segmental definition for /b/ is shown. Taking into account the global constraints for syllabification, assignment of associations and demarcation of phonological words, our example can be finally represented as seen at the end of (6).

(6)  `endtime := time:(start:Start & end:End & eigenperiod:EigenPeriod &`
           `release:Release) &`
        `equal(phase_point(Start, EigenPeriod, Release), End).`

---

[8] For clipping see (Kröger 1993a).

[9] Start and end of the gestural activation interval are *not* gestural control parameter, but result from our endpoint-based characterisation of temporal intervals.

[10] The adaption consists of exchanging the original reference to consonantal vs. vocalic gestures by reference to subsyllabic constituents.

[11] At the time of writing, our implementation still simulates arithmetic constraints by expressing arithmetic formulae with feature terms, which are handed over to a PROLOG derivate with built-in arithmetic constraint facilities for purposes of equation solving. However, Walther (1992) has already shown for a research prototype of CUF that a full integration by means of such a PROLOG derivate (CLP($\mathcal{R}$) or DM-CAI Sicstus PROLOG clone) is an unproblematic enterprise. Genuine integration of arithmetic constraints for the current CUF version is in preparation.

```
phase_point(Start, EigenPeriod, Phase) := add(Start,
        multiplicate(EigenPeriod, divide(Phase, 360))).

b := is(supralaryngeal_obstruent)
        & primary_articulation(lips & closed(thelips) & endtime
        & association_and_release_phase(stop)
        & obstruent_clipping(default))
        & voiced_in_onset & final_devoicing & theeigenperiod.
global_constraints := syllabify & is_phonological_word
        & association_rules.
ebbt := phon:[glottal_stop, ae, b, t, postphonatory_opening].

test(example) := ebbt & phon:global_constraints.
```

The constraint solving mechanism of CUF derives a complex structure when given the query **test(example)**, of which (7) shows the subpart of /b/[p] side by side with a visualisation of the gestural trajectories[12] and timing relationships as well as the oscillogram of **ebbt**.

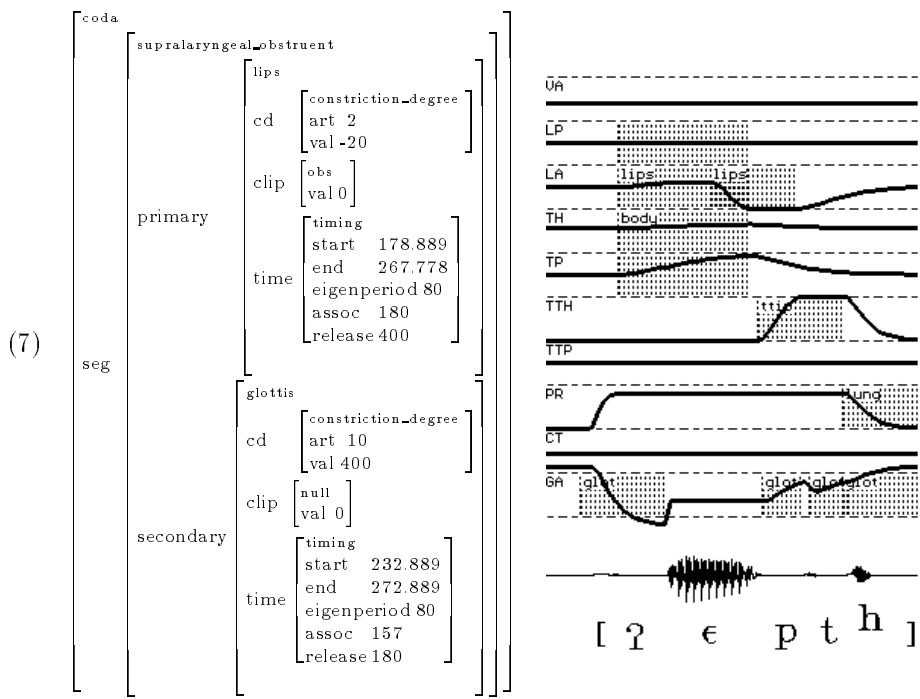

(7)

---

Importantly, the typed attribute-value structure in (7) contains *all* the information that is needed by the articulatory synthesizer in order to compute the sound event it denotes.[13] In particular, the missing explicit interface documents itself by the absence of e.g. an additional `phonetics` attribute together with a corresponding complex interpretational relationship between `phon` and `phonetics` values.

## 5  Discussion and summary

The finite-state and unification-based approaches of (Bird 1992) and (Bird & Klein 1993) share the concerns of this work in explicitly advancing the enrichment of HPSG by declaratively specified phonological information, but lack the quantitative-phonetic orientation pursued here.

In contrast, YORKTALK (Coleman 1992) constitutes a synthesis model within the DP paradigm which is explicit about phonetic realisation. Nonsegmental phonological descriptions and a large bank of 'phonetic exponency' specifications[14] serve to control a Klatt synthesizer in the *acoustic* domain. In our view, gestures as elegant bivalent primitives operating at the more natural *articulatory* level considerably simplify the task of phonetic realisation as seen from the perspective of phonology.

Similarly, the SYNPHONICS concept-to-speech system strives for phonetic interpretation within the framework of HPSG. Inter alia, an interesting direct link between semantic focus and phonetic accent realisation is discussed in (Günther 1993). Due to the planned employment of a Klatt synthesizer, this approach however shares problems of YORKTALK, while it seemingly also duplicates information by separating the phonological and phonetic domain in attribute-value structures. It remains unclear how exactly the authors of SYNPHONICS would translate between both domains, the latter of which is conceived of as being gestural in nature, too.

Without doubt the approach for coupling phonology and phonetics presented here is nothing but an initial step. On the one hand the study of gestural timing relationships under the influence of syllable structure has to be advanced. Initial research e.g. by Browman & Goldstein (1988) on principled global temporal organisation of whole syllable constituents in American English[15] could be verified for German by means of electromagnetic articulography. Such a study should then also incorporate medial clusters as well as consonantal nuclei. Also further linguistic dimensions such as word accent etc. deserve closer investigation as to their influence on timing. Machine learning of parametric time models from a gesturally segmented and annotated corpus should constitute a promising approach in this respect (cf. van Santen 1993).

---

[13] For purely technical reasons a trivial format conversion takes places with concomitant extraction of phonetically relevant attributes. Note that [art] values are simply numerical equivalents of articulatory types lips (2) and glottis (10), to be optimized away in future work. While [val(ue)] specifications give quantitative content to constriction degrees relative to articulator-specific scales, timing information is in milliseconds (apart from phase values).

[14] The latter have not been published apart from isolated examples, but are the main ingredient responsible for the quality of YORKTALK's speech output (p.c. Richard Ogden).

[15] According to their interpretation the arithmetic mean of temporal target locations of all onset gestures belonging to a syllable gets associated with the following vowel.

On the other hand the factual realisation of phonological processes must be studied and modelled. Does final devoicing possess a gradual nature, being moreover conditioned by pragmatic factors (Port & Crawford 1989)? Can German /ŋ/ realization (**Diphthong** [dɪftɔŋ] – **diphthongieren** [. . . ŋiːʁən]) be conceived of as intrasegmental variation of gestural phasing relationships? What about the syllable-dependent processes of /ʁ/-allophony (**frisch** [fʁɪʃ] – **Berg** [bɛɐ̯k] – **Bauer** [baʊɐ̯]), of ich/ach alternation combined with /g/ spirantisation[16] and schwa epenthesis? Can the latter be modelled by short, gesturally unspecified temporal gaps, as hypothesized in Walther (1992)? In conjunction with electromagnetic articulography experiments such hyphotheses can be evaluated very concretely in our approach.

The approach presented here addresses the problem of phonology–phonetics coupling by assuming a declarative constraint-based phonology component, which is realized in the constraint logic programming language CUF extended by arithmetic constraints (Walther 1992, 1993). In order to couple this component in a simple fashion and without explicit interfaces to the articulatory synthesizer of Kröger (1993 a,b), the central concept of gestures is employed. Initial practical results in synthesis demonstrate the general capabilities of our approach. Thus we may view our approach as a useful tool in future research directed towards an empirically anchored demarcation of linguistic phenomena: whether a given instance is best described as categorial-phonological or gradual-phonetic may be evaluated relative to an implemented model which can encompass both.

---

[16] Cf. Walther & Wiese (1993) for a formalisation of the latter two processes.